# Reasoning in Systems with Elements that Randomly Switch Characteristics

Subhash Kak

**Abstract.**
We examine the issue of stability of probability in reasoning about complex systems with uncertainty in structure. Normally, propositions are viewed as probability functions on an abstract random graph where it is implicitly assumed that the nodes of the graph have stable properties. But what if some of the nodes change their characteristics? This is a situation that cannot be covered by abstractions of either static or dynamic sets when these changes take place at regular intervals. We propose the use of sets with elements that change, and modular forms are proposed to account for one type of such change. An expression for the dependence of the mean on the probability of the switching elements has been determined. The system is also analyzed from the perspective of decision between different hypotheses. Such sets are likely to be of use in complex system queries and in analysis of surveys.

## Introduction

For reasoning in complex systems, one often uses static data structures designed for situations that do not change after they are constructed. Static sets allow only query operations on their elements — such as checking whether a given value is in the set, or enumerating the values in some arbitrary order. In dynamic sets, on the other hand, one can insert and delete elements. But what if the elements stay the same in number, yet their properties change randomly?

Consider a very basic model where the set of N objects is viewed as consisting of $n_A$ objects of Type A and the remaining of Type B, where the properties of the two types are distinct. The observation process consists of choosing a random element and examining it and then replacing it in the box. By repeated interaction with the system, the probability of A is easily computed:

$$P(A) = \frac{n_A}{N} \qquad (1)$$

Let the outcomes A and B be mapped to the random variable X with values 1 and -1, respectively. We can pick the threshold of 0 to determine which probability is greater. If $\sum X_i > 0$, then P(A) > P(B).

Now assume that the objects are actually of 3 types: A, B, and C with numbers that are $n_A$, $n_B$, and $n_C$, $n_A+n_B+n_C=N$, where C nodes are special in the sense that during query (interaction) they project the properties of A and B in some random fashion? One could also view the situation as consisting of $n_C$ objects of Type C that are non-classical or one could view Class C as a random set whose elements change between Type A and B.

An example of this will be polling surveys. Assume there are three political parties but only two candidates who belong to Parties A and B. The members of the third party must pick candidate A or B, where their choice varies from day to day based on news and public opinion (for simplicity, we assume here that members of the other two parties vote for their party candidates, for in reality there will be some cross-voting from the other groups as well).

This situation represents unstable sets with a subset of floating elements and this fact should be considered in determining system behavior.

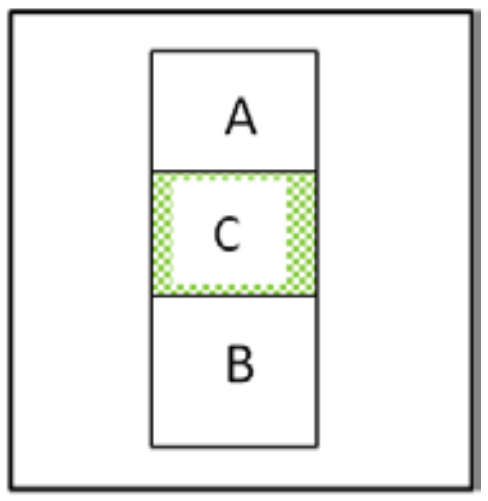

Figure1. Elements of C switch characteristics randomly between A and B

The probability P(A) will not stabilize as the system is queried (say on different days) and its value will fall in a range:

$$\frac{n_A}{N} \leq P(A) \leq \frac{n_A + n_C}{N} \qquad (2)$$

The C nodes may be seen as creating noise in the measurement and if indeed real systems behave like this this property needs to be taken into consideration and its impact on reasoning evaluated.

For purpose of illustration, assume $n_A$=5, $n_B$=2, and and $n_C$=3. If the transitions occur at the end of the day, then sampling of different systems will change the estimate of elements of Class A as in Figure 2 below.

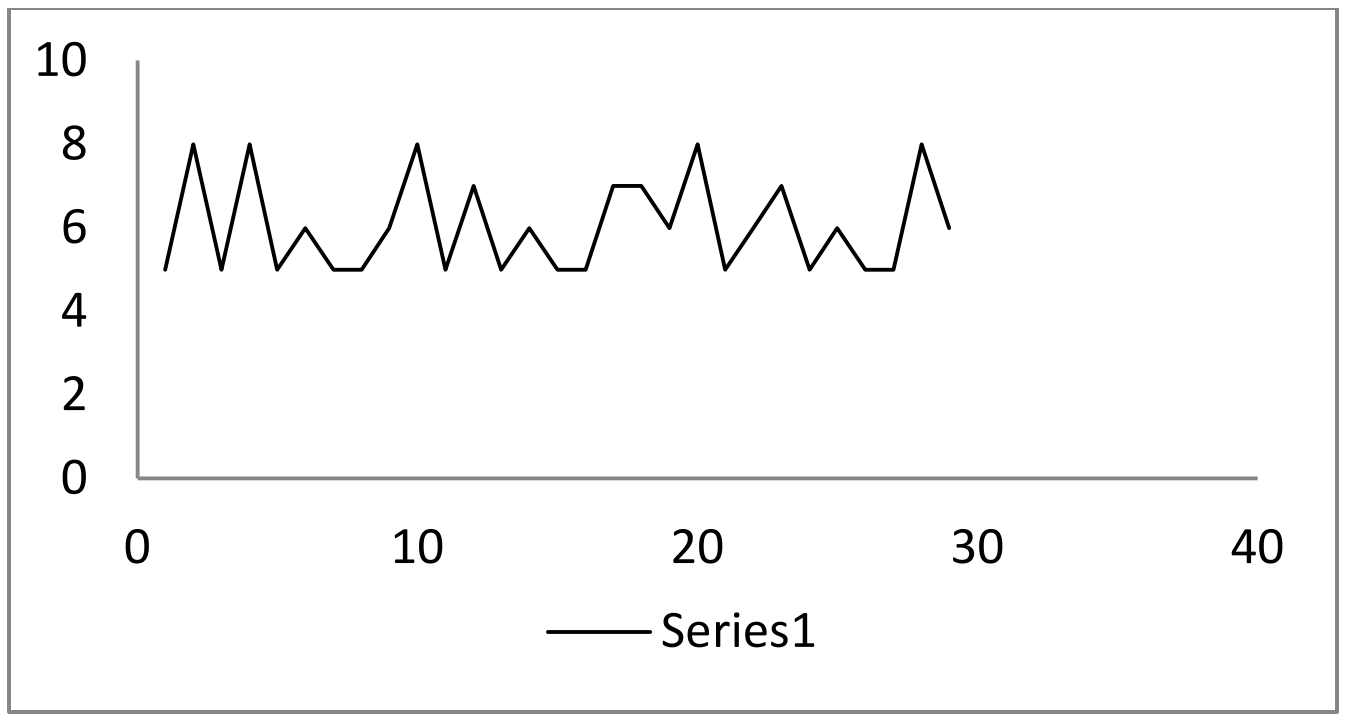

Figure 2. Changing count of the elements of A

As example of importance of such sets, consider that in medicine, a patient might respond to different active ingredients in different ways due to a certain element of interchangeability in the structure. The changeability will complicate analysis of Simpson's and related paradoxes [1][2]. The Simpson's paradox arises from the fact that if $a_1 / b_1 > c_1 / d_1$ and $a_2 / b_2 > c_2 / d_2$ there will be situations so that $(c_1 + c_2)/(d_1 + d_2) > (a_1 + a_2)/(b_1 + b_2)$, but if there is range associated with the numerator values, we get a more complicated picture related to the ranges where probability inversion occurs as seen in Table 1.

Table 1. Probability inversion

| | Week 1 | Week 2 | Total |
|---|---|---|---|
| Lisa | 0/1 | 3 or 4/5 | 3 or 4/6 |
| Bart | 1/4 | 2 or 3/3 | 3 or 4/7 |

Lisa and Bart are assigned the task of editing articles. In Week 2, their work gets further edited differently by supervisors. The performance of Bart is superior to Lisa's both weeks (excepting when Lisa has done 4/5 and Bart has done 2/3) but when the work is aggregated, Lisa is better.

Our consideration of interswitching may make it possible to do an indirect accounting of interaction networks that exist in the body and the complicating influence of information and regulatory networks to examine the placebo effect [3]. This effect is counterintuitive in that an inert substance turns out to be an effective treatment for some patients, and it arises out of the information flows and expectations associated with the treatment.

In some cases the interaction between elements may rightly be seen through the lens of quantum decision theory, especially when there is evidence of superpositional behavior by the nodes [4][5]. One major consequence of switching nodes in the quantum context is that probability measurements are not commutative. Thus the result P(T1,T2)≠P(T2,T1) even when T1 and T2 are apparently independent. The interaction disturbs the system that causes the variables to change, so that the order has an influence on the expectations associated with the system [6].

These ideas also apply in the field of belief vectors, which represent relations between states underlying the reasoning process. A specific belief may be a numerical value or a Boolean variable representing a choice between alternatives and in a basic form it may be seen as representing a class such as Type A. The choice itself might fluctuate between two different possibilities that may be viewed as being altered by the random influence projected by Class C. The dynamics of this projection may be such so that it does go to zero or some other fixed point.

In this paper, we consider the problem of reasoning with states of the system that change dynamically with each interaction. Specifically, we consider a model where the internal states change and a nonlinear mapping is proposed as a possible mechanism. Results on the statistical characteristics of the observations have been obtained. The system has also been analyzed from the perspective of noise theory.

**Dynamics of the system model**

Since the components of the system interact with each other, the system cannot be analyzed based on classical probability. But since the nature of these interactions is potentially endless, we must constrain our analysis by means of specific assumptions.

We assume the system to have nodes of type A, B, and C. The C nodes switch into A and B back and forth just like the population of the predators and prey in an isolated habitat. Such complex system dynamics will be characterized by fixed points and orbits. Different mechanisms may be at work in different systems.

Rather than speak of the full system, we will now focus only on the elements of Type C. Let the probability that an element of C will projected as A be *n*. Let each interaction with the system change the proportion of A and B nodes. Assume dynamics determined by the function s: N → N on the set of positive integers where 0 is taken to be an even number, and

$$s(i+1) = f(s(i)) \tag{3}$$

where *i* is the interaction count with the system.

Various modular schemes for the dynamic change of probability will be considered. Let *s*(*i*) be the number of Type C nodes that are observed as Class A nodes at *i*th measurement. Let the value of *i* run from 0 to M-1.

Scheme 1

$$s(i+1) = s(i) + k \bmod M \tag{4}$$

This will make the probability change in a zig-zag manner.

Scheme 2

$$s(i+1) = ks(i) \bmod M \tag{5}$$

If M is prime then the nodes will go through a cycle and the cycle is full if k is a primitive element of the prime.

Example 1. We have M=6, or a total of 5 nodes of Type C, and k = 2. The changes in the number of Type A nodes on each inspection will be as follows (where the first row represents the number in the first encounter):

| | | | | |
|---|---|---|---|---|
| 1 | 2 | 3 | 4 | 5 |
| 2 | 4 | 0 | 2 | 4 |
| 4 | 2 | 0 | 4 | 2 |

So in this case we have a fixed point (which is 0) and a cycle 2 4.

If the initial number of nodes was not 3, then the probability will switch between the values 4/5 and 2/5. For initial number 3, the probability goes to zero.

Scheme 3

Consider a general modular mapping:

$$s(n) = \begin{cases} \frac{n}{2}, & \text{if } n \text{ is even} \\ kn + 1 & \text{if } n \text{ is odd} \end{cases} \tag{6}$$

In particular, we consider this transformation for the specific case of k=3:

$$s(n) = \begin{cases} n/2 & \text{if } n \text{ is even} \\ 3n+1 \bmod M & \text{if } n \text{ is odd} \end{cases} \tag{7}$$

This is the finite version of the *3n+1* – mapping associated with the names of L. Collatz and other mathematicians [7][8]. The value of M determines the largest orbit that will be associated with the probability evolution. The initial count of nodes of Type A will be taken to be s(0). If M is even, the population of Type A nodes will never go down to zero excepting when *3i =1* mod M.

As example, consider the mapping for mod 17. The numbers 1 through 16 are mapped in one step as follows:

*s*(0) values →

| | | | | | | | | | | | | | | | |
|---|---|---|---|---|---|---|---|---|---|---|---|---|---|---|---|
| 1 | 2 | 3 | 4 | 5 | 6 | 7 | 8 | 9 | 10 | 11 | 12 | 13 | 14 | 15 | 16 |
| 4 | 1 | 10 | 2 | 16 | 3 | 5 | 4 | 11 | 5 | 0 | 6 | 6 | 7 | 12 | 8 |
| 2 | 4 | 5 | 1 | 8 | 10 | 16 | 2 | 0 | 16 | 0 | 3 | 3 | 5 | 6 | 4 |
| 1 | 2 | 16 | 4 | 4 | 5 | 8 | 1 | 0 | 8 | 0 | 10 | 10 | 16 | 3 | 2 |
| 4 | 1 | 8 | 2 | 2 | 16 | 4 | 4 | 0 | 4 | 0 | 5 | 5 | 8 | 10 | 1 |
| 2 | 4 | 4 | 1 | 1 | 8 | 2 | 2 | 0 | 2 | 0 | 16 | 16 | 4 | 5 | 4 |
| 1 | 2 | 2 | 4 | 4 | 4 | 1 | 1 | 0 | 1 | 0 | 8 | 8 | 2 | 16 | 2 |
| 4 | 1 | 1 | 2 | 2 | 2 | 4 | 4 | 0 | 4 | 0 | 4 | 4 | 1 | 8 | 1 |
| 2 | 4 | 4 | 1 | 1 | 1 | 2 | 2 | 0 | 2 | 0 | 2 | 2 | 1 | 4 | 4 |
| 1 | 2 | 2 | 4 | 4 | 4 | 1 | 1 | 0 | 1 | 0 | 1 | 1 | 4 | 2 | 2 |

(9)

Figure 2. The evolution of the initial states with each observation.
The fixed point is 0 and the orbit is 1 4 2 1

If M is odd, the Type A nodes will either have a fixed point or be part of an orbit.

*Example 2*. If *i(0)* = 7, and M= 55, one gets the sequence 7, 22, 11, 34, 17, 52, 26, 13, 40, 20, 10, 5, 16, 8, 4, 2, 1; in other words, the orbit or trajectory of 7 has 17 elements and the path length is 16. The maximum value generated is 52. Let N= 37, for the same initial distribution. We then get 7, 22, 11, 34, 17, 15, 9, 28, 14, 7, …. which is an orbit of 9.

The trajectory for *n* = 3 for M=100 is 3, 10, 5, 16, 8, 4, 2, 1, that is it has 8 elements in it or it has a path length of 7, with a maximum value of 16. The sequence for *n* = 25, takes 23 steps, climbing to 88 before descending to 1: ( 25, 76, 38, 19, 58, 29, 88, 44, 22, 11, 34, 17, 52, 26, 13, 40, 20, 10, 5, 16, 8, 4, 2, 1 ).

The assumption in the modular mappings above is that the system is classical although highly nonlinear. However there are situations where this is not true and the consideration of a quantum model appears justified [9][10[11][12]. This takes us beyond classical sets in a different manner.

**Statistical analysis**

Given that $n_A$, $n_B$, and $n_C$ are the number of elements belonging to the sets A, B, and C, respectively, and $N = (n_A + n_B + n_C)$, we can begin with the basic probability associated with events A and B.

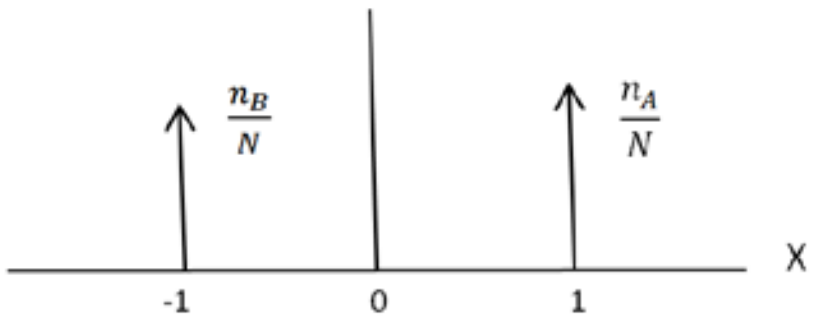

Figure 4. Probability of sets A and B

Let M be the variable representing the short-term average of X. Give the random switching of the elements of C, M takes values that range over $(n_A - n_B - n_C)/N$ to $(n_A - n_B + n_C)/N$. We will assume that M is a uniformly distributed random variable as shown below:

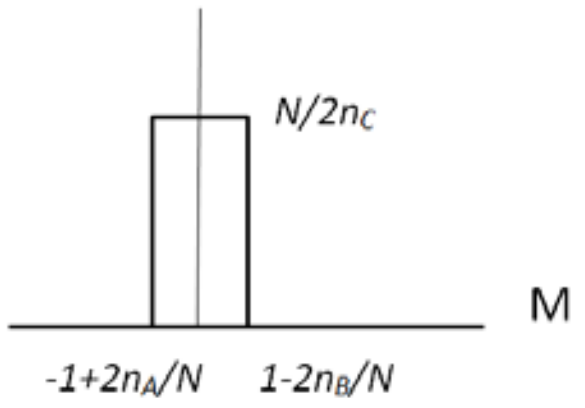

Figure 4. Distribution of M

The expected value of M, E(M)= E(X):

$$E(M) = \frac{(n_A - n_B)}{N} \quad (9)$$

The mean square value of M, E(M$^2$), may be easily computed:

$$E(M^2) = \frac{(3n_A^2 + 3n_B^2 + n_C^2 - 6n_A n_B)}{3N^2} \quad (10)$$

And the variance of M is

$$VAR(M) = \frac{n_C^2}{3N^2} \quad (11)$$

Thus the variance of the mean is one-third of the square of the proportion of switching elements. This variance is independent of the actual values of $n_A$ and $n_B$. An estimate of this variance will be a measure of the size of C, which may be important in determining how many individuals in the poll belong to a separate class where the choices are not firm.

**Class C as Noise**

We can also present a noise theory perspective on the problem. Assume, for the sake of symmetry, that Classes A and B have the same number of elements and that they represent two different Hypotheses A and B. A number of measurements, $X_i$, are taken and with A and B mapped to 1 and -1, the decision taken is by threshold logic:

Choose Hypothesis A if $\sum_i X_i \geq 0$; (12)
Otherwise, Hypothesis B.

Since we assume that the changes in the membership of C occurs with low rate (so that the surveys on a specific day, when the membership is stable) can be as large as one pleases. One might also assume now that elements of A and B can also change as in Figure 12 below. The source probabilities are expressed by subscript "s" and the observation probabilities are expressed by subscript "o".

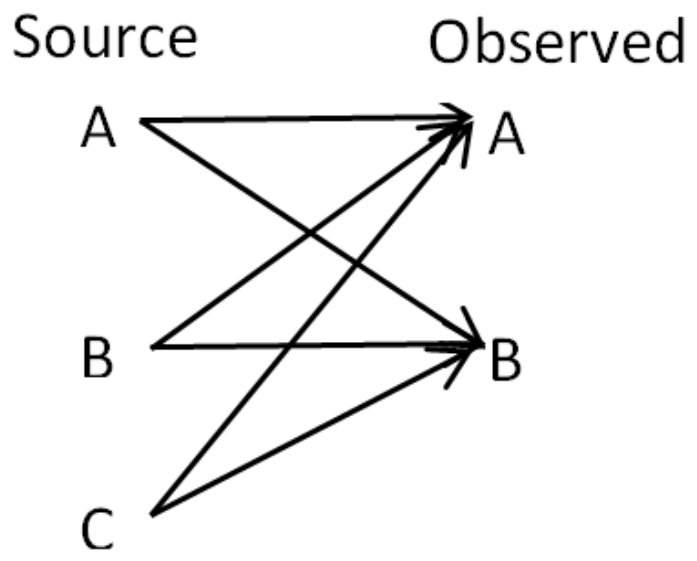

Figure 5. The observation process

We have

$$p(A_o) = p(A_o|A_s)p(A_s) + p(A_o|B)p(B_s) + p(A_o|C_s)p(C_s) \quad (13)$$

and so on.

By Bayes' theorem:

$$p(A_s|A_o) = \frac{p(A_o|A_s)p(A_s}{p(A_o|A_s)p(A_s)+p(A_o|B)p(B_s)+p(A_o|C_s)p(C_s)} \quad (14)$$

If the task is to estimate the size of A (or B), then Bayes' estimate may be used [13][14][15].

However, the random nature of the elements of set C implies that the probabilities $p(A_o|C)$and $p(B_o|C)$are not fixed. Under certain conditions, these probabilities may be taken to be fixed and constant. But in general, the nature of the nonlinear process underlying the change of the characteristics as well as the environment will determine what these probabilities are.

The query could be to ascertain the probability of various subsets associated with the system as a means of determining its structure. But as information about the subsets comes in, the sampling strategy will have to be changed to reflect the estimates obtained by Bayes' testing. This is a process that will have to be updated continuously.

**Conclusions**

We have considered sets where some of the elements switch characteristics in some random manner over time. This is a situation that cannot be covered by abstractions of either static or dynamic sets when these changes take place at regular intervals. We propose the use of sets with elements that change, and modular forms are proposed to account for one kind of such change. An expression for the dependence of the mean on the probability of the switching elements was determined. The system was also analyzed from the perspective of decision between different hypotheses. Such sets are likely to be of use in complex system queries and in analysis of surveys.